\pdfoutput=1

\documentclass[11pt]{article}

\usepackage[]{EMNLP2022}

\usepackage{times}
\usepackage{latexsym}
\usepackage{multirow}
\usepackage{array}
\usepackage{graphicx}
\usepackage{amssymb}
\usepackage{hhline}
\usepackage{amsmath}
\usepackage{cases}
\usepackage{mathtools}
\usepackage{enumitem}

\usepackage[T1]{fontenc}

\usepackage[utf8]{inputenc}

\usepackage{microtype}

\title{\textsc{ClidSum}: A Benchmark Dataset for Cross-Lingual \\ Dialogue Summarization}

\author{Jiaan Wang\textsuperscript{1,2}\thanks{ \ \ Work was done when Jiaan Wang was interning at Pattern Recognition Center, WeChat AI, Tencent Inc, China.}, \ Fandong Meng\textsuperscript{1}\thanks{ \ \ Corresponding authors.}, \ Ziyao Lu\textsuperscript{1}, \ Duo Zheng\textsuperscript{4}\\
\bf {Zhixu Li\textsuperscript{3}\footnotemark[2], \ Jianfeng Qu\textsuperscript{2} and \ Jie Zhou\textsuperscript{1}} \\
\small{\textsuperscript{1}Pattern Recognition Center, WeChat AI, Tencent Inc, China} \\
\small{\textsuperscript{2}School of Computer Science and Technology, Soochow University, Suzhou, China} \\
\small{\textsuperscript{3}Shanghai Key Laboratory of Data Science, School of Computer Science, Fudan University, Shanghai, China} \\
\small{\textsuperscript{4}Beijing University of Posts and Telecommunications, Beijing, China} \\
\small \texttt{jawang1@stu.suda.edu.cn}, \texttt{fandongmeng@tencent.com}, \texttt{zhixuli@fudan.edu.cn}
}

\begin{document}
\maketitle
\begin{abstract}

We present \textsc{ClidSum}, a benchmark dataset towards building cross-lingual summarization systems on dialogue documents.
It consists of 67k+ dialogue documents and 112k+ annotated summaries in different target languages.
Based on the proposed \textsc{ClidSum}, we introduce two benchmark settings for supervised and semi-supervised scenarios, respectively. 
We then build various baseline systems in different paradigms (pipeline and end-to-end) and conduct extensive experiments on \textsc{ClidSum} to provide deeper analyses.
Furthermore, we propose m\textsc{Dial}BART which extends mBART via further pre-training, where the multiple objectives help the pre-trained model capture the structural characteristics as well as key content in dialogues and the transformation from source to the target language.
Experimental results show the superiority of m\textsc{Dial}BART, as an end-to-end model, outperforms strong pipeline models on \textsc{ClidSum}.
Finally, we discuss specific challenges that current approaches faced with this task and give multiple promising directions for future research. We have released the dataset and code at \url{https://github.com/krystalan/ClidSum}.
\end{abstract}

\section{Introduction}
Cross-Lingual Summarization (XLS) aims at generating a summary in the target language from a given document in a different source language.
Existing XLS datasets mainly focus on single-participant
documents, such as news reports~\cite{zhu-etal-2019-ncls,nguyen-daume-iii-2019-global,bai-etal-2021-cross}, how-to guides~\cite{ladhak-etal-2020-wikilingua} and encyclopedia articles~\cite{perez-beltrachini-lapata-2021-models}.
Many efforts have been devoted to improving XLS for single-participant documents and achieved great success~\cite{yao-etal-2015-phrase,10.1109/TASLP.2018.2842432,ouyang-etal-2019-robust,cao-etal-2020-jointly,zhu-etal-2020-attend,maurya-etal-2021-zmbart,Chi2021mT6MP,liang-etal-2022-variational,wang2022survey}.

\begin{figure}[t]
\centerline{\includegraphics[width=0.45\textwidth]{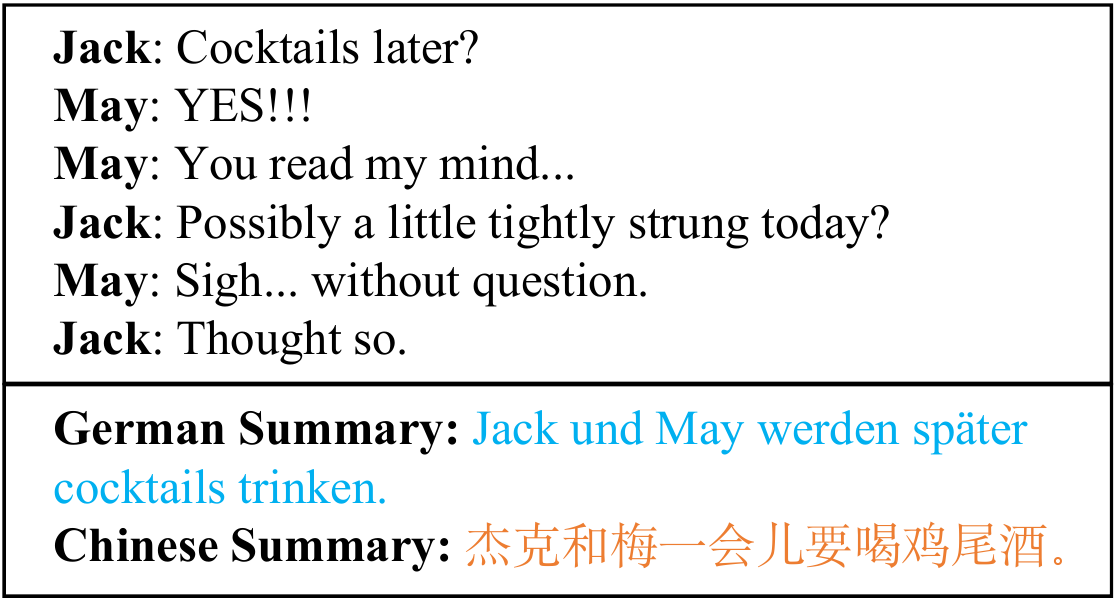}}
\caption{An example of cross-lingual dialogue summarization. \textcolor[RGB]{0,176,240}{Blue} and \textcolor[RGB]{237,125,49}{orange} sentences are the German and Chinese summaries of the given English dialogue, respectively.}
\label{fig:example}
\vspace{-0.5cm}
\end{figure} 

Different from single-participant documents, dialogue is a discourse produced by more than one person~\cite{Haviland1990LinguisticAL}.
The multi-participant dialogue documents, which record the communications between human and human/machine, have attracted wide research attention due to their key role in daily interpersonal interaction~\cite{Zhang2020DIALOGPTL}.
Meanwhile, the globalization progress has prompted conversations among interlocutors of different native languages and brought many scenarios, e.g., international academic conferences and business meetings.
Thus, it is valuable to provide speakers with the summary in their familiar language to help them efficiently grasp the gist of a foreign language dialogue.
Nevertheless, dialogue-oriented XLS is still under-explored due to the lack of corresponding datasets.

To this end, we introduce Cross-Lingual Dialogue Summarization (XLDS) task that aims to summarize a dialogue in the source language into a different language.
To promote the XLDS research, we construct \textsc{ClidSum} (\underline{\textbf{C}}ross-\underline{\textbf{LI}}ngual \underline{\textbf{D}}ialogue \underline{\textbf{SUM}}marization), the first large-scale XLDS benchmark dataset with three features: (1) The proposed \textsc{ClidSum} is based on two existing monolingual dialogue summarization datasets, i.e., SAMSum~\cite{gliwa-etal-2019-samsum} and MediaSum~\cite{zhu-etal-2021-mediasum}. We choose these two datasets under the consideration of 
their quality and diversity.
(2) To make these datasets suitable for XLDS, we employ professional translators to translate original English summaries of SAMSum and MediaSum to German and Chinese.
Eventually, the translated corpora constitute \textsc{ClidSum}, which totally contains \textasciitilde56.4k En$\Rightarrow$De and \textasciitilde56.4k En$\Rightarrow$Zh XLDS samples\footnote{En: English; De: German; Zh: Chinese.}.
(3) Besides the supervised benchmark setting that has been discussed by most previous XLS work, we argue that it is necessary to utilize large-scale monolingual dialogue summarization pairs for XLDS due to the dearth of cross-lingual samples. Thus, we design a semi-supervised setting, where a large number of monolingual pairs together with a relatively small scale of cross-lingual pairs are used to build XLDS systems.

Based on \textsc{ClidSum}, we build and evaluate various baseline systems, including \emph{summarize-then-translate}, \emph{translate-then-summarize} and \emph{end-to-end} paradigms:
(1) For \emph{summarize-then-translate} baselines, we use a monolingual dialogue summarizer to generate summaries and further translate them to the target language through a machine translation (MT) model/service.
(2) For \emph{translate-then-summarize} baselines, we utilize a MT model/service to translate the dialogue documents from the source language to the target language, and then obtain summaries via a monolingual dialogue summarizer.
(3) As for \emph{end-to-end} paradigm, we adopt mBART-50~\cite{Tang2020MultilingualTW}, a multi-lingual sequence-to-sequence (Seq2Seq) model which has been pre-trained on a large-scale corpora across 50 languages using denoising objectives.
Though such a general model achieves surprising results in many downstream tasks, its performance will naturally degrade when there is a relatively large gap between the pre-training and fune-tuning stages~\cite{lai-etal-2021-saliency-based,Zhong2021DialogLMPM}. 
Therefore, to narrow the gap and fully use large-scale monolingual dialogue summarization pairs provided in semi-supervised setting, we propose m\textsc{Dial}BART that extends mBART-50 through the second stage of pre-training with four objectives: action infilling, utterance permutation, monolingual dialogue summarization and machine translation.
Specifically, action infilling and utterance permutation encourage the model to effectively capture the structural characteristics in dialogues. Monolingual dialogue summarization gives the model the ability to summarize dialogues. Machine translation enables the model to learn the transformation from the source language to the target language.
The pre-training corpora are provided in the semi-supervised setting, which means we do not use additional data other than \textsc{ClidSum}.
The experimental results show that m\textsc{Dial}BART outperforms strong pipeline baselines on \textsc{ClidSum}.
We also conduct human studies to compare generated summaries from different methods and discuss specific challenges that current approaches faced with XLDS. We hope that our work could promote the development of XLDS.

Our main contributions are concluded as follows:
\begin{itemize}[leftmargin=*,topsep=0pt]
\setlength{\itemsep}{0pt}
\setlength{\parsep}{0pt}
\setlength{\parskip}{0pt}
    \item We introduce XLDS task and present \textsc{ClidSum}, the first large-scale XLDS benchmark dataset collecting about 56.4k En$\Rightarrow$De and 56.4k En$\Rightarrow$Zh samples via crowd-sourcing. Furthermore, we design supervised and semi-supervised benchmark settings on \textsc{ClidSum}.
    \item We elaborately build and evaluate various baselines of different paradigms (i.e., translate-then-summarize, summarize-then-translate and end-to-end) to provide deeper analyses.
    \item We propose the m\textsc{Dial}BART model that extends mBART-50 via the second pre-training stage. Experimental results show that m\textsc{Dial}BART outperforms existing baselines.
    \item To provide a deeper understanding of the XLDS task and the \textsc{ClidSum} dataset, we analyze the results of different methods and conclude the critical difficulties of XLDS on \textsc{ClidSum}.
\end{itemize}

\section{Related Work}

\label{subsec:xls}
\noindent \textbf{Cross-Lingual Summarization.} 
We divide existing XLS datasets into \textit{synthetic datasets} and \textit{multi-lingual website datasets} according to the construction methods.
(1) \textit{Synthetic datasets} are constructed through translating the summaries of existing text summarization datasets to a target language, such as En2ZhSum, Zh2EnSum~\cite{zhu-etal-2019-ncls} and En2DeSum~\cite{bai-etal-2021-cross}.
These datasets are further equipped with the round-trip translation strategy~\cite{zhu-etal-2019-ncls,zhang-etal-2021-crafting,lai-etal-2022-generating} to filter out low-quality samples.
(2) \textit{Multi-lingual website datasets} are collected from websites which provide multi-lingual versions for their articles. For instance, WikiLingual~\cite{ladhak-etal-2020-wikilingua} collects articles from the WikiHow website, where many English articles are translated to non-English versions by human writers. Each article also links to parallel articles in other languages, if available. Thus, it is handy to collect different language versions for one article.
Next, the summary of each language-specific article is extracted through a heuristic strategy.
In this way, the article in one language and its summary in a different language could constitute an XLS sample.
In the similar way, Global Voices~\cite{nguyen-daume-iii-2019-global} and XWikis~\cite{perez-beltrachini-lapata-2021-models} collect multi-lingual articles from Global Voices and Wikipedia websites, respectively.
Early XLS methods~\cite{Wan2010CrossLanguageDS,Zhang2016AbstractiveCS,Ayana2018ZeroShotCN,ouyang-etal-2019-robust} are based on pipeline paradigms due to the scarcity of parallel corpus.
Recently, \citet{zhu-etal-2019-ncls} propose the first large-scale XLS dataset and further explore the multi-task learning on end-to-end (e2e) XLS systems which achieve great improvement over pipeline methods.
Subsequently, many efforts have contributed to the e2e XLS systems.
Among them, \citet{zhu-etal-2020-attend} exploit the translation patterns in XLS. \citet{cao-etal-2020-jointly} propose a framework that jointly learns to align and summarize for XLS. \citet{Xu2020MixedLingualPF} explore pre-training strategy on XLS.
\citet{liang-etal-2022-variational} adopt conditional variational auto-encoder (CVAE)~\cite{Sohn2015LearningSO} to deal with XLS.
Different from existing XLS work, we shift research attention from single-participant documents to multi-participant dialogues.

\vspace{0.5ex}
\noindent \textbf{Dialogue Summarization.}
Dialogue summarization aims to summarize the dialogue document into a shorter passage. AMI~\cite{Carletta2005TheAM} and ICSI~\cite{Janin2003TheIM} are two early meeting corpora that both contain hundreds of samples. Recently, the SAMSum corpus~\cite{gliwa-etal-2019-samsum} is introduced, which contains over 16k chat conversations with human-labeled summaries.
MediaSum~\cite{zhu-etal-2021-mediasum} collects about 464k interview transcripts and uses the overview descriptions as summaries.
GupShup~\cite{mehnaz-etal-2021-gupshup} develops the first code-switched dialogue summarization dataset whose dialogues are in Hindi-English.
Based on these datasets, a lot of work~\cite{chen-yang-2020-multi,wu-etal-2021-controllable,Feng2021IncorporatingCK,chen-yang-2021-structure,feng-etal-2021-language} models the conversation characteristics and achieves great performance. 
All these efforts are given to monolingual or code-switched scenarios while we focus on cross-lingual scenario in this paper.

\begin{table*}[t]
  \centering
  \setlength{\belowcaptionskip}{-10pt}
  \resizebox{0.98\textwidth}{!}
  {
    \begin{tabular}{llcccccccc}
    \hline
    & \multicolumn{1}{c}{\multirow{2}{*}{\textbf{Dataset}}} & \multirow{2}{*}{\textbf{Domain}} & \multicolumn{1}{c}{\multirow{2}{*}{\textbf{Trans.}}} &  \textbf{Src}   & \textbf{Tgt}         & \multirow{2}{*}{\textbf{Documents}} & \multicolumn{1}{c}{\textbf{Doc.}}  & \multicolumn{1}{c}{\textbf{Src Summ.}} & \multicolumn{1}{c}{\textbf{Tgt Summ.}} \\
    & \multicolumn{1}{c}{}  & \multicolumn{1}{c}{} & \multicolumn{1}{c}{} & \textbf{Lang.} & \textbf{Lang.}       &  \multicolumn{1}{c}{}   & \multicolumn{1}{c}{\textbf{Length}} & \multicolumn{1}{c}{\textbf{Length}}     & \multicolumn{1}{c}{\textbf{Length}}     \\ \hline
    & En2ZhSum     & News & Auto & En         & Zh       & 370,687      &   755.0    &   55.2    & 96.0          \\
    & Zh2EnSum     & News & Auto & Zh         & En      & 1,699,713     & 103.7      & 17.9      & 13.7       \\
    & En2DeSum     & News & Auto & En         & De      & 437,797              &    31.0       & 8.5           & 7.5          \\ \hline
    \multirow{3}{*}{\rotatebox[origin=c]{90}{\textsc{\small{ClidSum}}} $\begin{dcases} \\ \\ \end{dcases}$}
    & XSAMSum      & Chat & Manual    & En         & De/Zh  & 16,369        & 83.9        & 20.3        &  19.9/33.0          \\
    & XMediaSum40k & Interview & Manual  & En         & De/Zh  & 40,000 &  1555.4    &    14.4         &   14.8/30.0          \\
    & MediaSum424k & Interview & -  & En         & -               & 423,596              &  1553.6           &      14.4       &   -           \\ \hline
    \end{tabular}
  }
  \caption{Statistics of \textsc{ClidSum} and previous synthetic XLS datasets. \textit{Trans.} indicates the translation method (automatic or manual) to construct dataset. \textit{Src Lang.} and \textit{Tgt Lang.} denote the source and target languages (En: English, Zh: Chinese or De: German) for each dataset. \textit{Documents} represents the size of each dataset. \textit{Doc. Length}, \textit{Src Summ Length} and \textit{Tgt Summ Length} show the average length of documents, source summaries and target summaries (word-level for English and German while character-level for Chinese) for each dataset, repectively.} 
  \label{table:statistic}
\end{table*}

\begin{table}[t]
  \centering
  \setlength{\belowcaptionskip}{-10pt}
  \resizebox{0.48\textwidth}{!}
  {
    \begin{tabular}{c|c|ccc}
    \hline
    \multicolumn{1}{c|}{\textbf{Dataset}}  & \textbf{Split} & \textbf{Dial.} & \textbf{Turns} & \textbf{Speakers} \\ \hline
    \multirow{3}{*}{(X)SAMSum}      & Train & 14,732&  11.2 &  2.4        \\
                                 & Dev.  & 818   &  10.8 &  2.4      \\
                                 & Test  & 819   &  11.2 &  2.4        \\ \hline
    \multirow{3}{*}{(X)MediaSum40k} & Train & 20,000&  30.4 &   9.3       \\
                                 & Dev.  & 10,000&  29.3 &   9.2       \\
                                 & Test  & 10,000&  30.6  &  9.3        \\ \hline
    \end{tabular}
  }
  \caption{Statistics of the dialogue characteristics in (X)SAMSum and (X)MediaSum40k. \textit{Dial.} indicates the number of dialogue documents for each subset of these datasets. \textit{Turns} and \textit{Speakers} represent the average number of utterances and speakers in dialogues.} 
  \label{table:dialogue_characteristics}
\end{table}

\section{\textsc{ClidSum}}
\label{sec:benchmark}

In this section, we first discuss how we choose existing monolingual dialogue summarization datasets for the construction of \textsc{ClidSum} (\S~\ref{subsec:data_collection}). Then, we introduce the details of the annotation process that translates original summaries to target languages (\S~\ref{subsec:data_annotation}). We next give statistic analysis of \textsc{ClidSum} (\S~\ref{subsec:data_analysis}). Finally, we formulate XLDS task (\S~\ref{subsec:task_definition}) and give the details of benchmark settings (\S~\ref{subsec:benchmark_settings}).

\subsection{Data Selection}
\label{subsec:data_collection}
As discussed in Section~\ref{subsec:xls}, there are two types of XLS datasets: \emph{synthetic datasets} and \emph{multi-lingual website datasets}. To our knowledge, there is no public website that provides multi-lingual dialogue-summary pairs. Therefore, we decide to construct XLDS dataset through translating original summaries of existing dialogue summarization datasets.
After carefully comparing existing datasets, we finally choose SAMSum~\cite{gliwa-etal-2019-samsum} and MediaSum~\cite{zhu-etal-2021-mediasum} datasets due to the following reasons:
(1) these datasets are of high quality, both of which consist of real-world or human-labeled monolingual dialogue-summary pairs;
(2) these datasets collect dialogues in a wide range of domains in daily life.

\subsection{Data Annotation}
\label{subsec:data_annotation}  

Since the size of MediaSum (\textasciitilde464k) is much larger than any other dialogue summarization datasets (typically less than 30k), it is costly to translate all summaries from MediaSum to target languages via crowd-sourcing.
Hence, we randomly select 20k samples from its training set, which together with all samples of validation set and testing set form MediaSum40k (totally 40k samples) subset. The remaining monolingual data is denoted as MediaSum424k.
We then translate original English summaries from SAMSum and MediaSum40k to German and Chinese through data annotation.

There are 55 professional translators, 3 data reviewers and 1 data expert participating in the annotation process.
All En$\Rightarrow$Zh translators have passed the TEM-8 while all En$\Rightarrow$De translators have passed both the TEM-8 and the PGH\footnote{\textit{Test for English Majors-Band 8} (TEM-8) and \textit{Prüfung für das Germanistik Hauptstudium} (PGH) are two language qualifications in China, which measure the overall English and German proficiency of undergraduates majoring in English and German, respectively.}.
The data reviewers and the expert are proficient in English, German and Chinese, and have rich experiences in checking translation quality.
During the annotation process, 20\% of the summaries translated by each translator are checked by a reviewer. If the accuracy\footnote{A translated summary will be regarded as incorrect once it has an error judged by humans.} is lower than 95\%, the translator needs to modify all his/her translated summaries under the guidance of the reviewer. To guarantee the overall quality, after the above process for one translation direction of SAMSum or MediaSum40k, 2\% of the translated summaries are randomly sampled and checked by the data expert. If the accuracy is lower than 95\%, the corresponding translators and reviewers need to revise their translated summaries again. This quality control loop is executed 7~\footnote{The accuracy achieves 90\% three times, 93\% once, 94\% twice and finally reaches 98\%.}, 1, 1 and 2 times for SAMSum(En$\Rightarrow$Zh), SAMSum(En$\Rightarrow$De), MediaSum40k(En$\Rightarrow$Zh) and MediaSum40k(En$\Rightarrow$De), respectively.

In the end, we collect 56,369 Chinese and 56,369 German summaries.
We denote the summary-translated SAMSum and MediaSum40k as XSAMSum and XMediaSum40k, respectively, which also inherit the data splitting of original datasets. The XSAMSum, XMediaSum40k and MediaSum424k together constitute \textsc{ClidSum} benchmark dataset.

\subsection{Data Analysis}
\label{subsec:data_analysis}
We compare \textsc{ClidSum} with previous \textit{synthetic datasets} in Table~\ref{table:statistic}.
\textsc{ClidSum} is the first large-scale XLS dataset towards dialogues, and it is also the first manually translated XLS dataset.
While one may argue that the scale of previous synthetic datasets is much larger than XSAMSum and XMediaSum40k, it is worth noting that 1) the scale of all dialogue summarization datasets is less than 30k except MediaSum, and 2) the quality of XSAMSum and XMediaSum40k is much higher than those automatically constructed ones.
In addition, as shown in Table~\ref{table:dialogue_characteristics}, the dialogues from (X)MediaSum40k contain more turns (\textasciitilde30 vs. \textasciitilde11) and more speakers (\textasciitilde9.3 vs. \textasciitilde2.4) than (X)SAMSum, leading to more challenges for XLDS research.

\subsection{Task Overview}
\label{subsec:task_definition}
Given a dialogue $\mathcal{D}=\{u_{1},u_{2},...,u_{|\mathcal{D}|}\}$ in the source language, where $u_{i}$ denotes the $i$-th utterance in $\mathcal{D}$, the XLDS task aims to generate corresponding summary $Y_{t} = (y_{1},y_{2},...,y_{|Y_{t}|})$ in a different target language, where $y_{j}$ denotes the $j$-th token in $Y_{t}$.

\subsection{Benchmark Settings}
\label{subsec:benchmark_settings}
We design two benchmark settings for supervised and semi-supervised scenarios, respectively. 
In the \textit{Supervised Setting}, an XLDS system is established on the training set of XSAMSum or XMediaSum40k and evaluated on the corresponding test set.
In the \textit{Semi-Supervised Setting}, the training set of XMediaSum40k and the whole MediaSum424k are used to build XLDS systems which are evaluated on the test set of XMediaSum40k.

\section{Baselines}

\subsection{Pipeline Method}
\label{subsec:pipeline}

The main idea of the pipeline method is decomposing XLDS task into dialogue summarization and machine translation sub-tasks. It can be further divided into \textit{summarize-then-translate} and \textit{translate-then-summarize} paradigms.

\noindent \textbf{Summarize-then-translate.} In this paradigm, a monolingual dialogue summarizer is used to generate summary in the same language with the given dialogue, and then obtain target language summary through machine translation.
The following models are adopted as the dialogue summarizers:
\begin{itemize}[leftmargin=*,topsep=0pt]
\setlength{\itemsep}{0pt}
\setlength{\parsep}{0pt}
\setlength{\parskip}{0pt}
    \item PEGASUS~\cite{Zhang2020PEGASUSPW} is a pre-trained abstractive summarization model.
    \item T5~\cite{Raffel2020ExploringTL} is a jointly pre-trained Seq2Seq model for many downstream NLP tasks.
    \item BART~\cite{Lewis2020BARTDS} is another pre-trained Seq2Seq model using denoising objectives during pre-training stage.
    \item mBART-50~\cite{Tang2020MultilingualTW} is a multi-lingual version of BART, which could be also employed on monolingual tasks, though it does not take advantage of its multi-lingual ability.
    \item MV-BART~\cite{chen-yang-2020-multi} is a BART-based multi-view dialogue summarization model which utilizes conversation information from different views, e.g., topic view and stage view.
    \item BART($\mathcal{D}_{\textsc{all}}$)~\cite{feng-etal-2021-language} encodes additional dialogue characteristics to BART. The characteristics are extracted by a DialoGPT~\cite{Zhang2020DIALOGPTL} annotator.
\end{itemize}

To translate the generated summaries from the source to the target language, previous XLS work~\cite{ladhak-etal-2020-wikilingua,perez-beltrachini-lapata-2021-models} only adopts sophisticated translation services, e.g., Google Translation and AWS Translation. Though great performances, the black-box APIs provided by translation services are constantly updating, which leads to poor reproducibility. Thus, we decide to adopt both translation services and models. Specifically, we consider three translation methods:
(1) Google Translation\footnote{\url{https://cloud.google.com/translate}};
(2) OPUS-MT~\cite{TiedemannThottingal:EAMT2020} releases lots of transformer-based MT models which are trained on OPUS corpus\footnote{\url{https://opus.nlpl.eu/}} with different translation directions;
(3) Trans-WMT20: we train two transformer-based Seq2Seq models on WMT20 parallel corpus\footnote{\url{https://www.statmt.org/wmt20/translation-task.html}} from scratch, including En$\Rightarrow$De and En$\Rightarrow$Zh.

\noindent \textbf{Translate-then-summarize.} This paradigm first translates English dialogues from \textsc{ClidSum} to German and Chinese through the same translation methods as the \textit{summarize-then-translate}, and then generate summaries via mBART-50.

\subsection{End-to-End Method}
\label{subsec:end2end}
The end-to-end method needs simultaneously learn both dialogue summarization and machine translation.
We finetune mBART-50 with input dialogues from the source language, and summaries from the target language.
Note that mBART-50 is used in both pipeline and end-to-end paradigms, where the languages of input and output sequences are the same when used in the pipeline paradigm while different in the end-to-end paradigm.
To indicate the input and output languages, mBART-50 appends language identifiers (e.g., \texttt{En}, \texttt{De} and \texttt{Zh}) at both the encoder and the decoder sides.

\begin{figure*}[t]
\centerline{\includegraphics[width=0.95\textwidth]{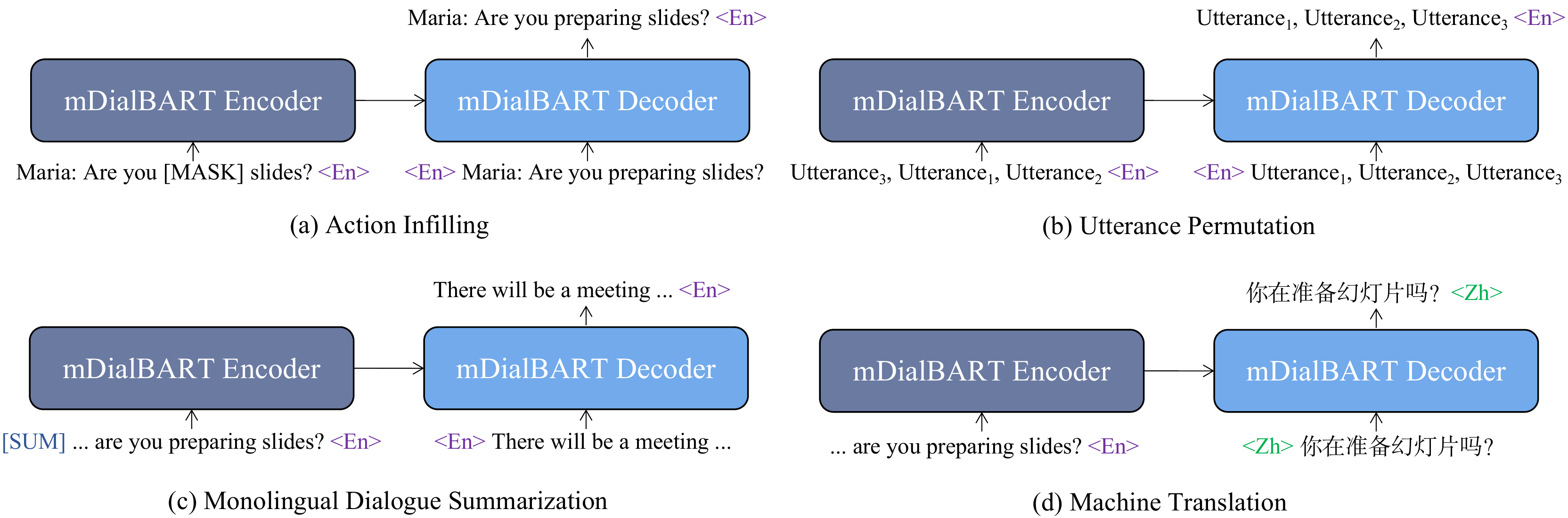}}
\caption{The second stage of pre-training in m\textsc{Dial}BART. \texttt{<En>} and \texttt{<Zh>} are two language identifiers indicating the input and output languages. \texttt{[SUM]} is a special token to indicate the summarization task.}
\label{fig:pretraining_task}
\end{figure*} 

\section{m\textsc{Dial}BART}  
\label{sec:mdialbart}
As suggested by previous work~\cite{zhu-etal-2020-attend,ladhak-etal-2020-wikilingua}, building an end-to-end model is preferable to a pipeline one due to: 1) the pipeline models suffer from error propagation; 2) the translation systems in pipeline paradigm require either a large parallel corpus to train MT models or the monetary cost of paid MT services; 3) the pipeline models have a recurring latency during inference~\cite{ladhak-etal-2020-wikilingua}.
Thus, it is valuable and urgent to explore end-to-end models on XLDS.
To this end, we propose m\textsc{Dial}BART which extends mBART-50 via a second stage of pre-training, where the objectives help the pre-trained model better adapt to the XLDS task (cf., Figure~\ref{fig:pretraining_task}).

\subsection{Pre-Training Tasks and Corpora}
\label{subsec:pretrain_obj}

\noindent \textbf{Action Infilling (AcI).}
Action triples (i.e., \textit{``who-doing-what''}) help the pre-trained model explicitly be aware of the actions within utterances for generating more factual summaries~\cite{chen-yang-2021-structure}.
Following \citet{chen-yang-2021-structure}, we extract action triples through \texttt{OpenIE} systems~\cite{angeli-etal-2015-leveraging}. Then we randomly sample text spans from action triples and replace each span with a \texttt{[MASK]} token. About 15\% of tokens in an utterance will be masked. The action infilling task requires the pre-trained model to reconstruct original dialogues based on the masked dialogues.

\vspace{0.5ex}
\noindent \textbf{Utterance Permutation (UP).} We shuffle all utterances from dialogues and encourage the model to restore them.

\vspace{0.5ex}
\noindent \textbf{Monolingual Dialogue Summarization (MDS).}
We adopt MDS as a pre-training task to enable the model to learn the summarization ability.

\vspace{0.5ex}
\noindent \textbf{Machine Translation (MT).}
Since the XLDS task needs translation capability, we employ machine translation as one of the pre-training tasks.

We pre-train m\textsc{Dial}BART on XMediaSum40k and MediaSum424k corpora which are provided by the semi-supervised setting of \textsc{ClidSum}. The dialogues as well as monolingual dialogue-summary pairs used in AcI, UP and MDS come from MediaSum424k. For MT, we use the original English paired with the translated summaries from the training set of XMediaSum40k as pre-training samples.

\subsection{Multi-Task Pre-Training}

For each iteration of the second pre-training stage, training samples from the four tasks are randomly selected and used to calculate the summation loss and update the parameters.
Thus, it is necessary to make the pre-trained model distinguish different tasks, to avoid the confusion brought by the joint training.
Specifically, AcI, UP and MT could be regarded as reconstruction tasks. To distinguish MDS from others, a special token \texttt{[SUM]} is appended at the beginning of input sequences of MDS.

\begin{figure}[t]
\centerline{\includegraphics[width=0.45\textwidth]{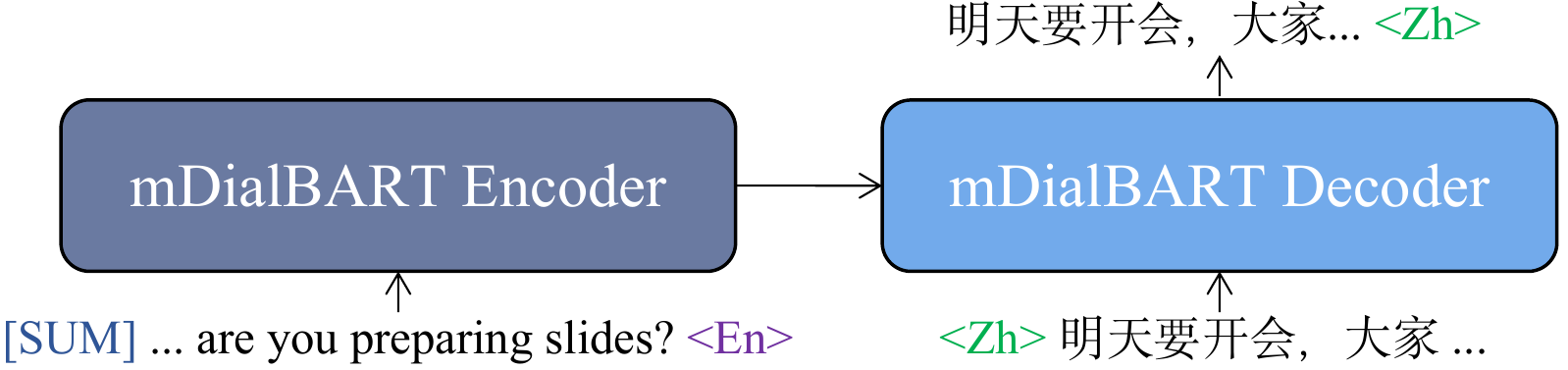}}
\caption{Finetune m\textsc{Dial}BART on XLDS.}
\label{fig:finetune}
\vspace{-0.4cm}
\end{figure}

\subsection{Fine-tuning on XLDS}
\label{subsec:finetune}
Figure~\ref{fig:finetune} shows the fine-tuning process on XLDS task. The input sequences of XLDS are prepended with token \texttt{[SUM]} to leverage dialogue summarization ability obtained from the second pre-training stage described in the previous section. Besides, the language identifiers at the encoder and the decoder sides are different to utilize the cross-lingual ability learned from the MT pre-training task.

\section{Experiments}

We evaluate m\textsc{Dial}BART and various baselines on \textsc{ClidSum}.
Four automatic metrics are adopted in our experiments: ROUGE-1 (R1) and ROUGE-2 (R2)~\cite{Lin2004ROUGEAP} evaluate unigram and bigram overlap between the generated summaries and correspondings references, respectively.
ROUGE-L (R-L)~\cite{Lin2004ROUGEAP} is applied to find the length of the longest common subsequence.
BERTScore (B-S)~\cite{Zhang2020BERTScoreET} evaluates the semantic similarity of generated sentences against the references.
For evaluation toolkits and the implementation details of all models, please refer to Appendix~\ref{appendix:implementation}.

\subsection{Main Results}
Table~\ref{table:main_experiment} shows the experimental results.
We first analyze the performance of pipeline baselines, and then compare them with end-to-end baseline. Second, we compare m\textsc{Dial}BART with all baselines to demonstrate its superiority. Lastly, we introduce a simple data augmentation strategy to further improve the performance of end-to-end methods.
For generated cases, please refer to Appendix~\ref{appendix:case}.

\begin{table*}[t]
  \centering
  \setlength{\belowcaptionskip}{-10pt}
  \resizebox{0.95\textwidth}{!}
  {
    \begin{tabular}{c|llcccc|cccc|cccc|cccc}
    \hline
\multirow{2}{*}{\#} & \multicolumn{1}{l}{}              & \multicolumn{1}{l|}{}              & \multicolumn{4}{c|}{XSAMSum$_{En\Rightarrow De}$} & \multicolumn{4}{c||}{XSAMSum$_{En\Rightarrow Zh}$} & \multicolumn{4}{c|}{XMediaSum40k$_{En\Rightarrow De}$} & \multicolumn{4}{c}{XMeidaSum40k$_{En\Rightarrow Zh}$} \\
& \multicolumn{1}{l}{}              & \multicolumn{1}{l|}{}              & \multicolumn{4}{c|}{R1 / R2 / R-L / B-S}   & \multicolumn{4}{c||}{R1 / R2 / R-L / B-S} & \multicolumn{4}{c|}{R1 / R2 / R-L / B-S}   & \multicolumn{4}{c}{R1 / R2 / R-L / B-S}  \\ \hline
    \multicolumn{19}{c}{Sum-Trans Methods}                                                      \\ \hline
1 & \multirow{3}{*}{PEGASUS}          & \multicolumn{1}{l|}{Trans-WMT}   & \multicolumn{4}{c|}{37.8 / 13.2 / 30.6 / 70.7}  & \multicolumn{4}{c||}{35.5 / 11.6 / 29.2 / 74.7} & \multicolumn{4}{c|}{19.8 / 6.5 / 17.5 / 59.0} & \multicolumn{4}{c}{21.6 / 5.8 / 18.9 / 64.8}  \\
2 & & \multicolumn{1}{l|}{OPUS-MT}       & \multicolumn{4}{c|}{41.5 / 16.7 / 34.0 / 73.4}   & \multicolumn{4}{c||}{33.3 / 11.0 / 27.8 / 72.1} & \multicolumn{4}{c|}{18.3 / 5.6 / 16.1 / 58.1} & \multicolumn{4}{c}{16.0 / 3.0 / 14.0 / 61.8}  \\
3 & & \multicolumn{1}{l|}{Google Trans.} & \multicolumn{4}{c|}{43.4 / 18.9 / 36.1 / 74.4} & \multicolumn{4}{c||}{38.5 / 14.3 / 31.9 / 75.8} & \multicolumn{4}{c|}{21.1 / 7.7 / 18.7 / 60.2} & \multicolumn{4}{c}{22.8 / 6.6 / 20.0 / 65.5} \\ \hline
4 & \multirow{3}{*}{T5}           & \multicolumn{1}{l|}{Trans-WMT}   & \multicolumn{4}{c|}{39.7 / 14.6 / 32.1 / 71.6}   & \multicolumn{4}{c||}{36.4 / 12.5 / 30.1 / 75.2} & \multicolumn{4}{c|}{18.6 / 6.3 / 16.0 / 56.5} & \multicolumn{4}{c}{21.7 / 5.7 / 18.5 / 64.4}  \\     
5 & & \multicolumn{1}{l|}{OPUS-MT}       & \multicolumn{4}{c|}{43.2 / 17.9 / 35.4 / 74.2}   & \multicolumn{4}{c||}{34.7 / 11.7 / 29.1 / 72.9} & \multicolumn{4}{c|}{17.5 / 5.4 / 15.1 / 57.0} & \multicolumn{4}{c}{17.0 / 3.3 / 14.6 / 61.5}  \\
6 & & \multicolumn{1}{l|}{Google Trans.} & \multicolumn{4}{c|}{45.3 / 20.3 / 37.4 / 75.2}   & \multicolumn{4}{c||}{39.7 / 15.2 / 33.0 / 76.6} & \multicolumn{4}{c|}{20.3 / 7.6 / 17.6 / 59.1} & \multicolumn{4}{c}{22.9 / 6.7 / 19.6 / 64.4}  \\\hline
7 & \multirow{3}{*}{BART}             & \multicolumn{1}{l|}{Trans-WMT}   & \multicolumn{4}{c|}{40.2 / 15.4 / 33.0 / 71.7}   & \multicolumn{4}{c||}{37.5 / 12.9 / 30.9 / 75.5} & \multicolumn{4}{c|}{20.0 / 6.7 / 17.7 / 59.3} & \multicolumn{4}{c}{23.4 / 7.0 / 20.6 / 65.6}  \\ 
8 & & \multicolumn{1}{l|}{OPUS-MT}       & \multicolumn{4}{c|}{44.3 / 19.2 / 36.7 / 74.7}   & \multicolumn{4}{c||}{36.7 / 12.7 / 30.5 / 73.6} & \multicolumn{4}{c|}{20.4 / 7.0 / 18.1 / 59.7} & \multicolumn{4}{c}{17.9 / 3.8 / 15.7 / 62.7}  \\
9 & & \multicolumn{1}{l|}{Google Trans.} & \multicolumn{4}{c|}{46.5 / 21.7 / 39.0 / 75.8}   & \multicolumn{4}{c||}{41.3 / 16.7 / 34.3 / 77.1} & \multicolumn{4}{c|}{23.7 / 9.9 / 21.1 / 61.9} & \multicolumn{4}{c}{24.7 / 8.0 / 21.9 / 66.3}  \\ \hline
10 & \multirow{3}{*}{mBART}     & \multicolumn{1}{l|}{Trans-WMT}   & \multicolumn{4}{c|}{39.1 / 14.0 / 31.3 / 70.9}   & \multicolumn{4}{c||}{36.5 / 12.1 / 30.1 / 75.0} & \multicolumn{4}{c|}{17.6 / 5.4 / 15.6 / 58.1} & \multicolumn{4}{c}{21.1 / 5.7 / 18.5 / 64.4}  \\      
11 & & \multicolumn{1}{l|}{OPUS-MT}       & \multicolumn{4}{c|}{41.3 / 16.2 / 33.5 / 73.0}   & \multicolumn{4}{c||}{32.1 / 10.3 / 27.0 / 71.6} & \multicolumn{4}{c|}{18.0 / 5.7 / 15.8 / 58.7} & \multicolumn{4}{c}{15.8 / 2.6 / 13.8 / 61.1}  \\
12 & & \multicolumn{1}{l|}{Google Trans.} & \multicolumn{4}{c|}{43.3 / 17.6 / 34.9 / 74.0}   & \multicolumn{4}{c||}{39.5 / 15.2 / 32.7 / 76.1} & \multicolumn{4}{c|}{20.8 / 7.9 / 18.4 / 60.5} & \multicolumn{4}{c}{22.6 / 6.7 / 19.8 / 65.1}  \\ \hline  
13 & \multirow{3}{*}{MV-BART}    & \multicolumn{1}{l|}{Trans-WMT}   & \multicolumn{4}{c|}{42.2 / 15.8 / 33.8 / 72.5}   & \multicolumn{4}{c||}{39.1 / 13.9 / 32.1 / 76.1} & \multicolumn{4}{c|}{20.5 / 6.7 / 17.9 / 59.5} & \multicolumn{4}{c}{23.7 / 7.7 / 20.6 / 65.9}   \\       
14 & & \multicolumn{1}{l|}{OPUS-MT}      & \multicolumn{4}{c|}{46.3 / 19.8 / 37.7 / 75.3}   & \multicolumn{4}{c||}{38.1 / 13.9 / 31.5 / 74.0} & \multicolumn{4}{c|}{20.9 / 7.4 / 18.7 / 60.2} & \multicolumn{4}{c}{18.5 / 3.8 / 16.2 / 62.7}   \\
15 & & \multicolumn{1}{l|}{Google Trans.} & \multicolumn{4}{c|}{\textbf{48.3} / \underline{22.6} / \underline{39.8} / \textbf{76.4}}   & \multicolumn{4}{c||}{\textbf{42.4} / \textbf{17.2} / \underline{34.9} / \underline{77.4}} & \multicolumn{4}{c|}{24.0 / 9.9 / 21.1 / 61.9} & \multicolumn{4}{c}{24.8 / 8.6 / 22.1 / 66.6}  \\ \hline
16 & \multirow{3}{*}{BART($\mathcal{D}_{\textsc{all}}$)}  & \multicolumn{1}{l|}{Trans-WMT}   & \multicolumn{4}{c|}{42.1 / 16.8 / 34.2 / 72.8}   & \multicolumn{4}{c||}{39.2 / 13.9 / 32.2 / 76.3} & \multicolumn{4}{c|}{20.4 / 6.7 / 17.9 / 59.6} & \multicolumn{4}{c}{23.6 / 7.6 / 20.8 / 66.0}  \\ 
17 & & \multicolumn{1}{l|}{OPUS-MT}       & \multicolumn{4}{c|}{45.3 / 19.2 / 37.0 / 74.7}   & \multicolumn{4}{c||}{35.9 / 13.2 / 30.2 / 73.8} & \multicolumn{4}{c|}{20.7 / 7.3 / 18.7 / 60.0} & \multicolumn{4}{c}{18.3 / 3.7 / 16.0 / 62.5}  \\
18 & & \multicolumn{1}{l|}{Google Trans.} & \multicolumn{4}{c|}{\underline{48.0} / \textbf{23.1} / \textbf{40.3} / \underline{76.3}}   & \multicolumn{4}{c||}{\underline{42.1} / \textbf{17.2} / \textbf{35.1} / \textbf{77.6}}  & \multicolumn{4}{c|}{23.9 / 9.9 / 21.2 / 62.0} & \multicolumn{4}{c}{24.8 / 8.6 / 22.2 / 66.8}  \\ \hline
\multicolumn{19}{c}{Trans-Sum Methods}                                                      \\ \hline
19 & \multicolumn{1}{l}{Trans-WMT}   & \multicolumn{1}{l|}{mBART}         & \multicolumn{4}{c|}{41.2 / 15.5 / 32.9 / 73.1}   & \multicolumn{4}{c||}{38.3 / 13.9 / 31.2 / 76.1}  & \multicolumn{4}{c|}{19.3 / 6.7 / 15.9 / 59.7} & \multicolumn{4}{c}{23.7 / 7.7 / 20.8 / 65.8}   \\
20 & \multicolumn{1}{l}{OPUS-MT}       & \multicolumn{1}{l|}{mBART}        & \multicolumn{4}{c|}{42.0 / 16.8 / 34.2 / 73.7}   & \multicolumn{4}{c||}{36.5 / 13.0 / 29.8 / 75.3}  & \multicolumn{4}{c|}{20.0 / 7.5 / 17.6 / 60.0} & \multicolumn{4}{c}{23.1 / 7.1 / 20.3 / 65.6}  \\
21 & \multicolumn{1}{l}{Google Trans.}  & \multicolumn{1}{l|}{mBART}         & \multicolumn{4}{c|}{43.5 / 17.8 / 35.1 / 74.1}   & \multicolumn{4}{c||}{40.0 / 14.9 / 32.6 / 76.6}  & \multicolumn{4}{c|}{20.9 / 8.2 / 18.5 / 60.4} & \multicolumn{4}{c}{24.1 / 8.2 / 21.4 / 65.9}  \\ \hline
\multicolumn{19}{c}{End-to-end Methods}                                                                       \\ \hline
22 & \multicolumn{2}{l|}{mBART}        & \multicolumn{4}{c|}{43.4 / 17.6 / 35.0 / 74.0}   & \multicolumn{4}{c||}{39.6 / 15.5 / 32.9 / 76.6} & \multicolumn{4}{c|}{20.6 / 7.7 / 18.2 / 60.4} & \multicolumn{4}{c}{23.8 / 7.8 / 21.0 / 66.0}   \\
23 & \multicolumn{2}{l|}{\quad+DA$^{\dagger}$}    & \multicolumn{4}{c|}{- / - / - / -}   & \multicolumn{4}{c||}{- / - / - / -} & \multicolumn{4}{c|}{23.1 / 9.8 / 20.6 / 61.8} & \multicolumn{4}{c}{25.9 / 9.2 / 23.2 / 66.9}   \\
24 & \multicolumn{2}{l|}{m\textsc{Dial}BART (our)$^{\dagger}$}    & \multicolumn{4}{c|}{- / - / - / -}   & \multicolumn{4}{c||}{- / - / - / -} & \multicolumn{4}{c|}{\underline{24.9} / \underline{10.3} / \underline{22.2} / \underline{62.8}} & \multicolumn{4}{c}{\underline{27.4} / \underline{10.5} / \underline{24.5} / \underline{68.0}}   \\
25 & \multicolumn{2}{l|}{\quad+DA$^{\dagger}$}    & \multicolumn{4}{c|}{- / - / - / -}   & \multicolumn{4}{c||}{- / - / - / -} & \multicolumn{4}{c|}{\textbf{26.7} / \textbf{12.1} / \textbf{24.0} / \textbf{63.8}} & \multicolumn{4}{c}{\textbf{28.7} / \textbf{11.1} / \textbf{25.7} / \textbf{68.4}}   \\ \hline
\end{tabular}
}
\caption{Experimental results on \textsc{ClidSum}. The \textbf{bold} and \underline{underline} denote the best and the second scores, respectively. ${\dagger}$ represents that the evaluated models utilize the monolingual dialogue-summary pairs provided by the semi-supervised setting. Sum-Trans: Summarize-then-translate, Trans-Sum: Translate-then-summarize.}
\label{table:main_experiment}
\end{table*}

\begin{table}[t]
  \centering
  \setlength{\belowcaptionskip}{-10pt}
  \resizebox{0.45\textwidth}{!}
  {
    \begin{tabular}{l|cc|cc}
    \hline
    \multicolumn{1}{c|}{\multirow{2}{*}{Methods}} & \multicolumn{2}{c|}{En$\Rightarrow$De} & \multicolumn{2}{c}{En$\Rightarrow$Zh} \\
    \multicolumn{1}{c|}{}                         & BLEU $\uparrow$         & TER $\downarrow$          & BLEU $\uparrow$        & TER $\downarrow$        \\ \hline

    Trans-WMT                                     & 25.7         & 74.4         & \underline{35.5}         & \underline{55.7}        \\
    OPUS-MT                                       & \underline{27.3}         & \underline{59.9}         & 24.1         & 70.2        \\
    Google Trans.                                 & \textbf{40.4}         & \textbf{47.9}         & \textbf{47.8}         & \textbf{44.4}        \\ \hline
    \end{tabular}
  }
  \caption{Translation results on WMT20 (En$\Rightarrow$De and En$\Rightarrow$Zh \textit{newstest2020}). $\uparrow$ indicates higher is better. $\downarrow$ indicates lower is better.} \vspace{-0.2cm}
  \label{table:mt_res}
\end{table}

\vspace{0.5ex}
\noindent \textbf{Pipeline Baselines.} Comparing the performance between translate-then-summarize (Trans-Sum) and summarize-then-translate (Sum-Trans), we find that Trans-Sum outperforms Sum-Trans accompanied with the same summarizer (row 19-21 vs. row 10-12) due to the limited amount of monolingual dialogue-summary pairs in the source language. Besides, the English-centric dialogue summarization work~\cite{chen-yang-2020-multi,feng-etal-2021-language} reduces the summarization error and helps Sum-Trans overtake Trans-Sum (row 15/18 vs. row 21).
Moreover, as shown in Table~\ref{table:mt_res}, we test the performance of the adopted MT methods on \textit{WMT20-newstest2020}. Google Trans performs best in both En$\Rightarrow$De and En$\Rightarrow$Zh directions, and OPUS-MT outperforms Trans-WMT in En$\Rightarrow$De translation, but is worse in En$\Rightarrow$Zh translation. With the same dialogue summarizer, the XLDS performance of pipeline methods is consistent with the performance of the corresponding MT method.

\vspace{0.5ex}
\noindent \textbf{End-to-End vs. Pipeline Baselines.} Comparing the performance between end-to-end and pipeline baselines, mBART-50 achieves competitive results with the strong Trans-Sum baselines (row 22 vs. row 21), but performs worse than the strong Sum-Trans baselines (row 22 vs. row 15/18).
This is because: (1) the strong pipeline baselines adopt the sophisticated translation service which leverages a large amount of parallel data; (2) the end-to-end models need both the abilities to translate and summarize, which requires a large amount of cross-lingual training data. Nevertheless, existing datasets can not fully meet this requirement.

\vspace{0.5ex}
\noindent \textbf{m\textsc{Dial}BART vs. All Baselines.} To further explore the end-to-end paradigm on XLDS, we propose m\textsc{Dial}BART that extends mBART-50 through the second pre-training stage.
Experimental results show that m\textsc{Dial}BART outperforms all baselines in all automatic metrics (row 24).
Note that the data used in the second pre-training stage is provided in the semi-supervised setting of \textsc{ClidSum}. That is, m\textsc{Dial}BART is a semi-supervised XLDS system and thus we only evaluate m\textsc{Dial}BART on the test set of XMediaSum40k.

\vspace{0.5ex}
\noindent \textbf{Data Augmentation.} Moreover, we construct a large number of pseudo-XLDS samples by translating original English summaries from MediaSum424k to German and Chinese via Google Translation.
The pseudo-XLDS samples together with the real XLDS samples from XMediaSum40k jointly train end-to-end models via a simple curriculum learning~\cite{Bengio2009CurriculumL} strategy, i.e., the training data is converted from pseudo data to real data in each epoch.
With the help of the pseudo-XLDS samples, mBART-50 and m\textsc{Dial}BART significantly improve their performance on XMediaSum40k (row 23/25 vs. row 22/24).
We also find that our m\textsc{Dial}BART outperforms mBART+DA (row 24 vs. row 23), indicating the effectiveness of the second stage pre-training.
Through carefully designed pre-training objectives, the pre-trained model is more effective than simply using the data augmentation strategy.

\begin{table}[t]
  \centering
  \setlength{\belowcaptionskip}{-10pt}
  \resizebox{0.45\textwidth}{!}
  {
    \begin{tabular}{l|c|c}
    \hline
    \multicolumn{1}{c|}{\multirow{2}{*}{Model}} & \multicolumn{1}{c|}{En$\Rightarrow$De} & \multicolumn{1}{c}{En$\Rightarrow$Zh} \\
    \multicolumn{1}{c|}{}    & R1 / R2 / R-L / B-S   & R1 / R2 / R-L / B-S       \\ \hline
    mBART     & 20.6 / 7.7 / 18.2 / 60.4         & 23.8 / 7.8 / 21.0 / 66.0     \\
    \quad w/ AcI       & 21.8 / 8.4 / 19.3 / 61.1         & 24.9 / 8.4 / 22.0 / 66.4         \\
    \quad w/ UP        & 21.4 / 8.0 / 18.8 / 60.8         & 24.2 / 8.1 / 21.4 / 66.1         \\
    \quad w/ MDS       & \underline{23.3} / \underline{9.3} / \underline{20.7} / \underline{62.1}         & \underline{26.0} / \underline{9.0} / \underline{23.0} / \underline{67.1}         \\
    \quad w/ MT        &  21.1 / 7.9 / 18.6 / 60.6        & 24.3 / 8.4 / 21.5 / 66.2         \\ \hline
    mDialBART & \textbf{24.9} / \textbf{10.3} / \textbf{22.2} / \textbf{62.8}        & \textbf{27.4} / \textbf{10.5} / \textbf{24.5} / \textbf{68.0}         \\ \hline
    \end{tabular}
  }
  \setlength{\belowcaptionskip}{3pt}
  \caption{Ablation study on XMediaSum40k.} \vspace{-0.2cm}
  \label{table:ablation}
\end{table}

\begin{table}[t]
  \centering
  \setlength{\belowcaptionskip}{-10pt}
  \resizebox{0.40\textwidth}{!}
  {
    \begin{tabular}{l|r|r|r}
    \hline
    \multicolumn{1}{c|}{Model} & \multicolumn{1}{c|}{Gram.}                       & \multicolumn{1}{c|}{Info.}                       & \multicolumn{1}{c}{Conci.}                      \\ \hline
    BART($\mathcal{D}_{\textsc{all}}$)+Google      & 0.240                          &  0.117                           &  0.283                      \\
    Google+mBART     &  -0.254                            &           -0.126                     &   -0.343                     \\
    mBART             & -0.334                               & -0.277                                &             -0.300           \\
    m\textsc{Dial}BART           & \textbf{0.348}                         &    \textbf{0.286}                                   & \textbf{0.360}                       \\ \hline
    \end{tabular}

  }
  \caption{Human studies on XMediaSum40k (En$\Rightarrow$Zh).} \vspace{-0.2cm}
  \label{table:human_study}
\end{table}

\subsection{Ablation Study}

As shown in Table~\ref{table:ablation}, we conduct ablation studies on XMediaSum40k to evaluate the contribution of each pre-training task.
All four tasks contribute to the second pre-training stage.
The most important one is MDS, which is most relevant to XLDS in all pre-training tasks.
Both MDS and XLDS need the pre-trained model to understand the main content of dialogues and further summarize them.
MT and UP bring less benefits than others due to the fewer pre-training samples and the lower required capability, respectively. Specifically, the number of MT pre-training samples is 20k, which is significantly fewer than other tasks.
UP only requires the pre-trained model to restore the order of utterances in noisy dialogues rather than predicting new words. Such NLU-style task would make it easier for models to learn shortcuts rather than the semantic understanding and reasoning~\cite{du-etal-2021-towards}.
For ablation studies of combining two or three tasks, please refer to Appendix~\ref{appendix:ablations}.

\subsection{Human Study}
We conduct human studies to further evaluate the performances of the strong baselines under each paradigm and our pre-trained model, i.e., BART($\mathcal{D}_{\textsc{all}}$) + Google Trans, Google Trans + mBART, mBART and m\textsc{Dial}BART.
We randomly select 50 samples from the test set of XMediaSum40k. Seven crowd workers with high levels of fluency in English and Chinese are asked to assess the generated Chinese summaries from three aspects: \textit{grammaticality} (Gram.), \textit{informativeness} (Info.) and \textit{conciseness} (Conci.).
Following the Best-Worst Scaling method~\cite{kiritchenko-mohammad-2017-best}, crowd workers are asked to select the best and the worst generated summaries on each criteria.
The result scores are calculated based on the percentage of times each model is selected as best minus the times it is selected as worst. Thus, the final scores should range from -1 (worst) to 1 (best).
Table~\ref{table:human_study} shows the results of human studies. Our m\textsc{Dial}BART outperforms strong baselines in all metrics, indicating its superiority.
The Fleiss' Kappa scores~\cite{fleiss1971measuring} of Gram., Info. and Conci. are 0.37, 0.26 and 0.43, respectively, indicating a good inter-agreement among our evaluators.

\subsection{XLDS Difficulties}
\label{subsec:xlds_diff}
To further study the specific challenges of end-to-end XLDS and give multiple promising directions for future research, we take a closer look at \textsc{ClidSum} and model generation errors.
We conclude the following difficulties worthy of research attention:

\vspace{0.5ex}
\noindent \textbf{Multiple Topics.} The dialogues in MediaSum40k are interview transcripts from NPR and CNN~\cite{zhu-etal-2021-mediasum}, and each transcript usually records multiple topics from different events.
Previous dialogue summarization work~\cite{chen-yang-2020-multi,feng-etal-2021-language} has proved the effectiveness of topic information and proposed topic-aware summarizers.
Thus, it is also necessary to explore topic-aware end-to-end XLDS models.

\vspace{0.5ex}
\noindent \textbf{Low Resource.} End-to-end XLDS models learn both dialogue summarization and machine translation simultaneously, which require a large amount of training data.
Intuitively, as a more difficult task, XLDS needs more training samples than MT when building an end-to-end model.
Following previous MT work~\cite{lin-etal-2020-pre}, a parallel corpus is considered as an extremely low resource when the number of its samples is less than 100k. Therefore, it is hard to learn XLDS well when only utilizing supervised data provided by \textsc{ClidSum}. We believe the low-resource XLDS is also worth studying.

\vspace{0.5ex}
\noindent \textbf{Domain Adaption.} There are various dialogue domains in daily life (e.g., chat, interview and debates), and it is impractical to construct XLDS datasets for each domain. Therefore, it is valuable to utilize XLDS data of one domain to improve the model performance of others.
Moreover, we find that m\textsc{Dial}BART performs slightly worse than mBART baseline when fine-tuning on XSAMSum, indicating its limited domain adaption capability.

\section{Conclusion}
In this paper, we introduce XLDS task and present \textsc{ClidSum}, the first large-scale XLDS benchmark dataset.
We also propose m\textsc{Dial}BART, a XLDS-oriented pre-trained model that extends mBART-50 via the second pre-training stage.
The carefully designed pre-training tasks help m\textsc{Dial}BART better adapt to the XLDS task.
Experiments on \textsc{ClidSum} demonstrate that our m\textsc{Dial}BART outperforms strong baseline models.

\section*{Ethical Considerations}
We discuss the main ethical considerations of \textsc{ClidSum} benchmark dataset as follows:
(1) Licenses. \textsc{ClidSum} is derived from the SAMSum~\cite{gliwa-etal-2019-samsum} and the MediaSum~\cite{zhu-etal-2021-mediasum}, both of which are well-constructed and published datasets. 
We will follow the  CC BY-NC-ND 4.0 license of SAMSum to make XSAMSum public.
Following the requirements of MediaSum, we restrict our usage to research purpose only, and will release the translated MediaSum under CC-BY-NC-SA 4.0 license.
(2) Compensation. During the translation annotation, the salary for annotating each summary is determined by the average time of annotation and local labor compensation standard.
(3) Data characteristics. We refer readers to the content and \cite{gliwa-etal-2019-samsum,zhu-etal-2021-mediasum} for more detailed characteristics.
(4) Potential problems. While principled measures are taken to ensure the quality of the dataset, there might still be potential problems with the dataset quality, which may lead to incorrect translations in applications.
We also consider potential ethical issues of our m\textsc{Dial}BART pre-trained model: m\textsc{Dial}BART inherits mBART-50~\cite{Tang2020MultilingualTW} and is further trained on the XMediaSum40k and MediaSum424k corpora.  Therefore, m\textsc{Dial}BART could reflect the same biases and toxic behaviors exhibited by language models, such as biases about race and gender~\cite{sheng-etal-2020-towards}.

\section*{Limitations}
While we show that m\textsc{Dial}BART performs best on XMediaSum40k, there are limitations that provide avenues for future work.
(1) As we discussed in Section~\ref{subsec:xlds_diff}, the domain adaption capability of m\textsc{Dial}BART is limited. We think future work can focus on designing general or unified XLDS pre-trained models which could be applied in multiple dialogue domains.
(2) Not all the parameters of m\textsc{DialBART} are completely useful, such as the token embeddings of irrelevant languages, which reduces the inference speed.

\section*{Acknowledgements}
We would like to thank anonymous reviewers for their suggestions and comments. This work is supported by the National Natural Science Foundation of China (No.62072323, 62102276), Shanghai Science and Technology Innovation Action Plan (No. 22511104700), the Natural Science Foundation of Jiangsu Province (Grant No. BK20210705), and the Natural Science Foundation of Educational Commission of Jiangsu Province, China (Grant No. 21KJD520005).

\bibliography{anthology}
\bibliographystyle{acl_natbib}

\appendix

\begin{figure*}[t]
\centerline{\includegraphics[width=0.80\textwidth]{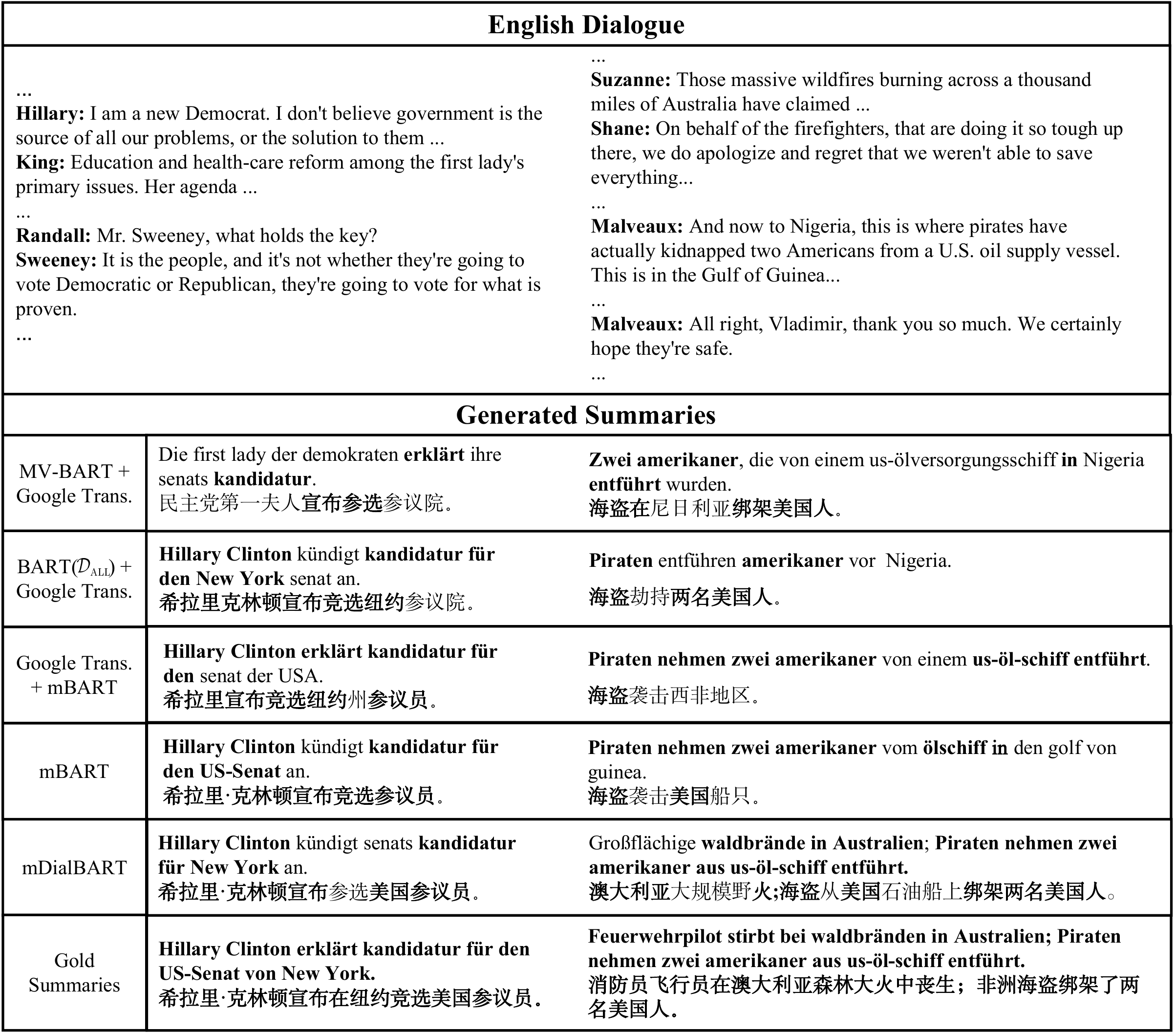}}
\caption{Example English dialogue in XMediaSum40k and the corresponding generated summaries of strong baselines and m\textsc{Dial}BART. The \textbf{bold} indicates the overlap words between generated summaries and gold summaries.}
\label{fig:case_study}
\vspace{-0.2cm}
\end{figure*} 

\section{Implementation Details}
\label{appendix:implementation}

\subsection{Automatic Evaluation}
\label{appendix:automatic_evaluation}
To calculate ROUGE scores, we employ the \textit{multi-lingual ROUGE} toolkit\footnote{\url{https://github.com/csebuetnlp/xl-sum/tree/master/multilingual_rouge_scoring}} that considers segmentation and stemming algorithms for various languages.
To calculate BERTScore, we use the \textit{bert-score} toolkit\footnote{\url{https://github.com/Tiiiger/bert_score}. The Chinese and German BERTScores are calculated through \textit{chinese-bert-wwm-ext} and \textit{bert-base-german-uncased} models, respectively.}.

\subsection{Translation Models}
For OPUS-MT models, we directly utilize the pre-trained \textit{OPUS-MT-en-de}\footnote{\url{https://huggingface.co/Helsinki-NLP/opus-mt-en-de}} and \textit{OPUS-MT-en-zh}\footnote{\url{https://huggingface.co/Helsinki-NLP/opus-mt-en-zh}} models.
For Trans-WMT translation model, we follow~\citet{liang-etal-2021-modeling} to train transformer-base models~\cite{vaswani2017attention} (512 hidden size, 2048 filter size, 8 multi-head attention, 6 encoder layers and 6 decoder layers) on WMT20 parallel corpus\footnote{\url{https://www.statmt.org/wmt20/translation-task.html}}.
Specifically, for En$\Rightarrow$De translation, we combine six corpora including Euporal, ParaCrawl, CommonCrawl, TildeRapid, NewsCommentary, and WikiMatrix.
And for En$\Rightarrow$Zh translation, we combine News Commentary v15, Wiki Titles v2, UN Parallel Corpus V1.0, CCMT Corpus, and WikiMatrix.
To pre-process the raw data, we employ a series of open-source/in-house scripts, including full-/half-width conversion, unicode conversation, punctuation normalization, and tokenization~\cite{wang-etal-2020-tencent}.
After filtering steps, we generate subwords via joint BPE~\cite{sennrich-etal-2016-neural} with 32K merge operations. Finally, we obtain 45,541,367 sentence pairs for En$\Rightarrow$De and 22,244,006 sentence pairs for En$\Rightarrow$Zh, respectively.

\subsection{Pre-Trained Language Models}
The implementation of pre-trained models used in our baselines is provided by the Huggingface Transformers~\cite{Wolf2020TransformersSN}, i.e., \textit{PEGASUS-large}\footnote{\url{https://huggingface.co/google/pegasus-large}}, \textit{T5-large}\footnote{\url{https://huggingface.co/t5-large}}, \textit{BART-large}\footnote{\url{https://huggingface.co/facebook/bart-large}} and \textit{mBART-50}\footnote{\url{https://huggingface.co/facebook/mbart-large-50-many-to-many-mmt}\label{mbart50}}.
In the fine-tuning process of pipeline baselines, we set the batch size of PEGASUS, T5, BART and mBART to 16 for XSAMSum and 24 for XMediaSum40k. The corresponding learning rates are set to 4e-5 and 2e-5, respectively. All these models are fine-tuned with 20 epochs. The hyperparameters of MV-BART~\cite{chen-yang-2020-multi} and BART($\mathcal{D}_{\textsc{all}}$)~\cite{feng-etal-2021-language} are the same as original paper.
In the fine-tuning process of end-to-end baseline, mBART-50 is fine-tuned with 4 batch size, 5e-6 learning rate and 20 epochs.

\subsection{m\textsc{Dial}BART}
Our m\textsc{Dial}BART first inherits from the \textit{mBART-50}$\textsuperscript{\ref{mbart50}}$ (1024 hidden size, 16 multi-head attention, 12 encoder layers and 12 decoder layers) and then suffers from the second stage of pre-training, which are conducted utilizing 8 NVIDIA Tesla V100 GPUs with 32GB memory.
m\textsc{Dial}BART is trained using \textit{pytorch-lightning}\footnote{\url{https://github.com/PyTorchLightning/pytorch-lightning}} framework with 5e-6 learning rate and 8 batch size.
%
We set the warmup steps and total steps to 5,000 and 1,000,000 steps, respectively.
The pre-trained m\textsc{Dial}BART is next fine-tuned on the XLDS task using a single GPU with 4 batch size and 5e-6 learning rate. The maximum number of tokens for input sequences is 1024.
In the test process, the beam size is 5, and the maximum decoded length is 150.

\section{Case Study}
\label{appendix:case}
We give the generated summaries of strong baselines and m\textsc{Dial}BART in Figure~\ref{fig:case_study}.
The dialogue in the left example only discusses one event, i.e., Hillary Clinton declares candidacy for the U.S. Senate seat from New York. All these models can generate a suitable summary that matches this event.
However, the content of the right dialogue across from many events, we find that all baselines capture only one event (i.e., the pirate event) while our m\textsc{Dial}BART is aware of the two different events in the dialogue and exactly summarize it.

\begin{table}[t]
  \centering
  \setlength{\belowcaptionskip}{-10pt}
  \resizebox{0.50\textwidth}{!}
  {
    \begin{tabular}{c|l|c|c}
    \hline
    \multicolumn{1}{c|}{\multirow{2}{*}{\#}} & \multicolumn{1}{c|}{\multirow{2}{*}{Model}} & \multicolumn{1}{c|}{En$\Rightarrow$De} & \multicolumn{1}{c}{En$\Rightarrow$Zh} \\
    \multicolumn{1}{c|}{}  & \multicolumn{1}{c|}{}   & R1 / R2 / R-L / B-S   & R1 / R2 / R-L / B-S       \\ \hline
    1 & mBART     & 20.6 / 7.7 / 18.2 / 60.4         & 23.8 / 7.8 / 21.0 / 66.0     \\ \hline
    2 & \quad +AcI       & 21.8 / 8.4 / 19.3 / 61.1         & 24.9 / 8.4 / 22.0 / 66.4         \\
    3 & \quad +UP       &  21.4 / 8.0 / 18.8 / 60.8     & 24.2 / 8.1 / 21.4 / 66.1   \\
    4 & \quad +MDS        & 23.3 / 9.3 / 20.7 / 62.1      & 26.0 / 9.0 / 23.0 / 67.1   \\
    5 & \quad +MT        & 21.1 / 7.9 / 18.6 / 60.6      & 24.3 / 8.4 / 21.5 / 66.2    \\ \hline
    6 & \quad +AcI+UP        & 22.1 / 8.5 / 19.6 / 61.3      & 25.1 / 8.5 / 22.2 / 66.6    \\
    7 & \quad +AcI+MDS        & 24.0 / 9.4 / 21.2 / 62.3       & 26.2 / 9.2 / 23.5 / 67.6   \\ 
    8 & \quad +AcI+MT        & 22.8 / 9.4 / 20.3 / 61.6      & 26.0 / 9.8 / 23.1 / 66.8   \\ 
    9 & \quad +UP+MDS        & 23.8 / 9.4 / 21.1 / 62.3      & 26.6 / 9.3 / 23.6 / 67.6   \\ 
    10 & \quad +UP+MT        & 22.2 / 9.1 / 19.8 / 61.2      & 25.3 / 9.7 / 22.6 / 66.5   \\ 
    11 & \quad +MDS+MT        & 24.2 / 10.2 / 21.6 / 62.4      & 27.2 / 10.2 / 24.2 / 67.6   \\ \hline
    12 & \quad +AcI+UP+MDS        & 23.7 / 9.7 / 21.1 / 62.2      & 26.6 / 9.4 / 23.5 / 67.5   \\
    13 & \quad +AcI+UP+MT        &  22.9 / 9.5 / 20.4 / 61.6     & 26.0 / 10.0 / 23.1 / 66.7   \\ 
    14 & \quad +AcI+MDS+MT        & \textbf{24.9} / \textbf{10.5} / \textbf{22.2} / \textbf{62.9}      & \textbf{27.7} / 10.5 / \textbf{24.7} / 67.9   \\ 
    15 & \quad +UP+MDS+MT        & 24.8 / \textbf{10.5} / 22.1 / 62.7      & 27.4 / \textbf{10.6} / 24.5 / 67.9   \\ \hline
    16 & mDialBART &  \textbf{24.9} / 10.3 / \textbf{22.2} / 62.8        & 27.4 / 10.5 / 24.5 / \textbf{68.0}         \\ \hline
    \end{tabular}
  }
  \caption{Ablation study of m\textsc{Dial}BART.} 
  \label{table:full_ablations}
  \vspace{-0.2cm}
\end{table}

\section{Effect of Each Pre-Training Task}
\label{appendix:ablations}
Table~\ref{table:full_ablations} shows the complete ablation studies of m\textsc{Dial}BART on XMediaSum40k.
We find that combining two tasks in the second pre-training stage outperforms using only one of them, e.g., mBART (+AcI+UP) outperforms both mBART (+AcI) and mBART (+UP). mBART (+MDS+MT) performs best in the setting of combining two tasks due to the high relevance\footnote{XLDS could be regarded as the combination of MDS and MT.} with XLDS.

As for combining three or four tasks, we find that MDS and MT are vital to the second pre-training stage. The performance of the pre-trained model with both MDS and MT is significantly better than others (row 14-16 vs. row 12-13).
In addition, the performance of mBART (+AcI+MDS+MT), mBART (+UP+MDS+MT) and m\textsc{Dial}BART is similar.
We think this is because both AcI and UP help the pre-trained model capture the structural characteristics in dialogues and these two tasks are redundant to some extent.

\end{document}